\documentclass[conference,a4paper]{IEEEtran}
\ifCLASSINFOpdf
\else
\fi
\usepackage{times}
\usepackage{epsfig}
\usepackage{graphicx}
\usepackage{amsmath}
\usepackage{amssymb}

\usepackage{rotating}

\begin{document}
%
\title{Log-Likelihood Score Level Fusion for Improved Cross-Sensor Smartphone Periocular Recognition}



%
\author{\IEEEauthorblockN{Fernando Alonso-Fernandez\IEEEauthorrefmark{1},
Kiran B. Raja \IEEEauthorrefmark{2}, Christoph Busch
\IEEEauthorrefmark{2}, Josef Bigun \IEEEauthorrefmark{1}}
\IEEEauthorblockA{\IEEEauthorrefmark{1}Halmstad University. Box 823. SE 301-18 Halmstad, Sweden\\
Email: \{feralo, josef.bigun\}@hh.se}
\IEEEauthorblockA{\IEEEauthorrefmark{2} Norwegian University of
Science and Technology, Gj\o{}vik, Norway\\
Email: \{kiran.raja, christoph.busch\}@ntnu.no}}


\maketitle

\begin{abstract}
The proliferation of cameras and personal devices results in a wide
variability of imaging conditions, producing large intra-class
variations and a significant performance drop when images from
heterogeneous environments are compared.
However, many applications require to deal with data from different
sources regularly, thus needing to overcome these interoperability
problems.
Here, we employ fusion of several comparators to improve periocular
performance when images from different smartphones are compared.
We use a probabilistic fusion framework based on linear logistic
regression, in which fused scores tend to be log-likelihood ratios,
obtaining a reduction in cross-sensor EER of up to 40\% due to the
fusion.
Our framework also provides an elegant and simple solution to handle
signals from different devices, since same-sensor and cross-sensor
score distributions are aligned and mapped to a common probabilistic
domain. This allows the use of Bayes thresholds for optimal decision
making, eliminating the need of sensor-specific thresholds, which is
essential in operational conditions because the threshold setting
critically determines the accuracy of the authentication process in
many applications.
\end{abstract}


%
\IEEEpeerreviewmaketitle

\section{Introduction}

The periocular region, the area surrounding the eye, has shown a
surprisingly high discrimination ability, while requiring the least
constrained acquisition among ocular or facial modalities
\cite{[Nigam15]}. It has thus become a very popular modality due to
the proliferation of unconstrained or uncooperative scenarios, e.g.
surveillance or smartphones \cite{[Alonso16]}. However, this massive
availability of devices results in heterogeneous quality between
probe and gallery images, which is known to reduce performance
significantly when different capture devices are used
\cite{[Jillela14]}. Even if the sensors work in the same spectrum,
they may have different spatial sampling rate, illumination sources,
field of view, etc. thus resulting in a challenge of
interoperability despite operating in the same spectrum
\cite{[Jain16]}.

This paper evaluates the fusion of different recognition systems to
improve cross-sensor recognition of images from different
smartphones. We use five periocular comparators based on popular
features from the literature, and the Visible Spectrum Smartphone
Iris (VSSIRIS) database \cite{[Raja14b]}, containing images from two
smartphones.
The individual comparators provide accurate recognition when
comparing images from the same device (with EER$\sim$0\%), but a 4-
to 10-fold EER increase is observed if images are not from the same
device.
%
%
There is also correlation between their performance and the size of
extracted templates. While the most accurate comparator provides
$\sim$0\% EER, it has a template size and comparison time that might
be prohibitive for real-time recognition in devices with limited
processing capabilities.
Fusion improves cross-sensor EER in more than 40\%, demonstrating
the validity of the proposed approach.
We employ a trained fusion based on linear logistic regression
\cite{[Alonso10]}, in which scores are mapped to
log-likelihood-ratios.
As a result, scores are in the same probabilistic,
sensor-independent domain, regardless whether they come from
comparison trials from same-sensor or different-sensor images,
greatly simplifying the fusion process. 

The rest of the paper is as follows. The periocular comparators
employed are described in Section II. Section III describes the
database and experimental protocol. Results of individual
comparators and fusion experiments are presented in Sections IV and
V, respectively. Conclusions are given in Section VI.

\section{Periocular Recognition Systems}

This section describes the five machine experts evaluated. \\


\begin{figure}[t]
\centering
\includegraphics[width=0.41\textwidth]{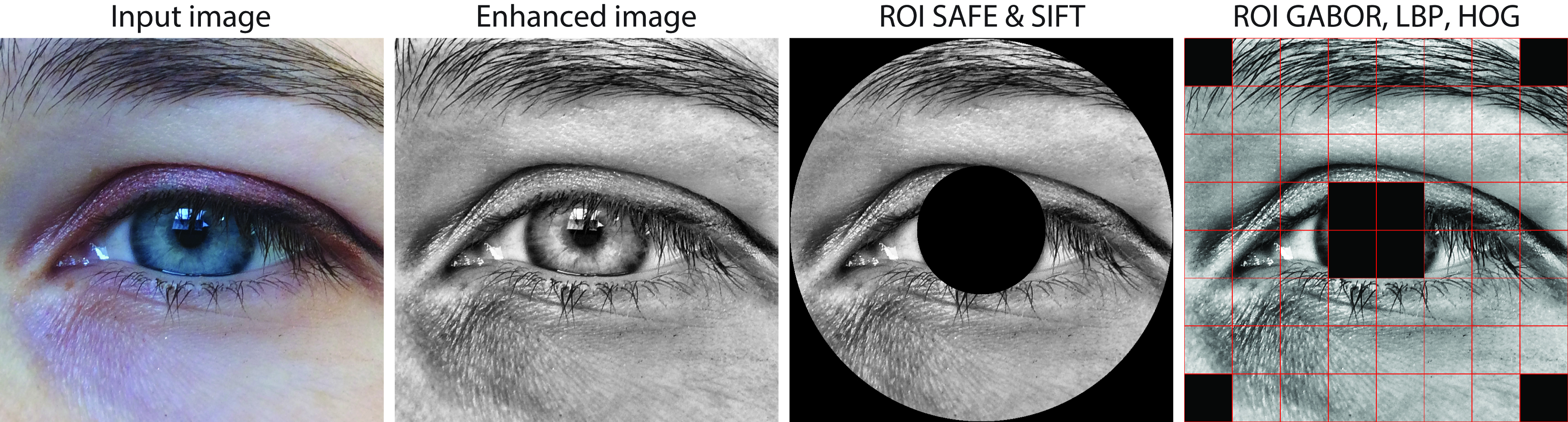}
\caption{Example image from VSSIRIS database. First/second columns:
input/preprocessed image with CLAHE. Third: ROI of SAFE and SIFT
comparators. Fourth: ROI of GABOR, LBP and HOG comparators (for
consistency with SAFE/SIFT, center and corner blocks are
discarded).} \label{fig:img-ROI}
\end{figure}

\noindent \textbf{Symmetry Patterns} based on the Symmetry
Assessment by Feature Expansion \textbf{(SAFE)} descriptor
\cite{[Alonso16a]}, which encodes the presence of various symmetric
curve families in concentric annular rings around image key-points.
We use the sclera center as unique key-point. The system employs 6
different scales for feature extraction, with 3 disjoint rings and 9
symmetry families per scale.
The first annular ring starts at the sclera circle, and the last
ends at the image boundary.
The ROI is shown in Figure~\ref{fig:img-ROI} (third column).
The sclera is used as anchor point, both to compute the eye center
and to estimate the ROI, due to its invariance to iris dilation. 
\\

\noindent \textbf{Gabor Features (GABOR)}.
The image is decomposed into non-overlapped blocks
(Figure~\ref{fig:img-ROI}, fourth column), and the local power
spectrum is then sampled at the center of each block by a set of
Gabor filters organized in 5 frequency and 6 orientation channels
\cite{[Alonso15]}.
This sparseness of the sampling grid allows direct filtering in the
image domain without needing the Fourier transform, with significant
computational savings.
\\

\noindent \textbf{SIFT key-points (SIFT)}
\cite{[Lowe04]}
%
with the adaptations described in \cite{[Alonso09]} for iris images,
particularly a post-processing step to remove
spurious keypoints using geometric constraints.\\

\noindent \textbf{Local Binary Patterns (LBP) and Histogram of
Oriented Gradients (HOG)}. Together with SIFT key-points, LBP
\cite{[Ojala02]} and HOG \cite{[Dalal05]} are the most widely used
features in periocular research \cite{[Alonso16]}.
The image is decomposed into non-overlapped regions
(Figure~\ref{fig:img-ROI}).
%
%
Then, HOG and LBP features are extracted from each block, quantized
into 8 different values (8 bins histogram) per block, with
histograms further normalized to account for
local illumination and contrast variations. 

\begin{figure}[htb]
     \centering
     \includegraphics[width=.4\textwidth]{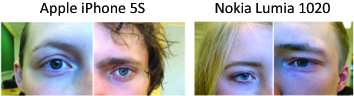}
     \caption{Sample images from VSSIRIS database (taken from \cite{[Raja14b]}).}
     \label{fig:db}
\end{figure}

\section{Database and Experimental Protocol}
\label{sect:db-protocol}

We use the Visible Spectrum Smartphone Iris (VSSIRIS) database
\cite{[Raja14b]}, having 28 semi-cooperative subjects (56 eyes)
captured indoors with
two smartphones (iPhone 5S and Nokia Lumia 1020, with images of
3264$\times$2448 and 3072$\times$1728 pixels, respectively), without
flash.
%
%
Each eye has 5 samples per smartphone, so
5$\times$56=280 images per device are available.
Figure~\ref{fig:db} shows some examples.
All images are annotated manually, so radius and center of the
iris circles are available. 
%
Images are resized by bicubic interpolation to have the same sclera
radius ($R$=145, average of the database), then they are aligned by
extracting a region of $6R$$\times$$6R$ (871$\times$871) around the
sclera center. This size is set empirically to ensure that all
images have sufficient margin to the four sides.
We use the sclera for normalization since it is not affected by
dilation.
Images are further equalized with CLAHE \cite{[Zuiderveld94clahe]}
to compensate local illumination variability 
%
(Figure~\ref{fig:img-ROI}).

We carry out verification experiments, comparing images both from
the same device (\emph{same-sensor}) and different devices
(\emph{cross-sensor}). Each eye is considered a different instance.
Genuine comparison trials are done by comparing each image of an
instance to the remaining images of the same eye, avoiding symmetric
comparisons.
This results in 10$\times$56=560 (same-sensor) and
5$\times$5$\times$56=1400 (cross-sensor) scores per smartphone.
Impostor trials are done by comparing the 1$^{st}$ image of an
instance to the 2$^{nd}$ image of the remaining eyes, resulting in
56$\times$55=3080 scores both in same- and
cross-sensor tests. 
%
Experiments have been done in a Dell E7240 laptop (i7-4600
processor, 16 Gb DDR3 RAM, built-in Intel HD Graphics 4400) with MS
Windows 8.1 Pro. The algorithms are implemented in Matlab r2009b
x64, with the exception of SIFT that is in
C++\footnote{http://vision.ucla.edu/$\sim$vedaldi/code/sift/assets/sift/index.html}
and invoked from Matlab via MEX files.
%
Size of stored template files and the extraction and matching
computation times are given in Table~\ref{tab:file-size-time}.

%
%
%
%
%
%
%
%

%
%
%
%
%
%
%

\begin{table}[htb]
\small
\begin{center}
\begin{tabular}{cccc}


\multicolumn{1}{c}{\textbf{}} &
 \multicolumn{1}{c}{\textbf{Template}} &
 \multicolumn{1}{c}{\textbf{Extraction}} &
  \multicolumn{1}{c}{\textbf{Comparison}}  \\

\multicolumn{1}{c}{\textbf{system}} &
 \multicolumn{1}{c}{\textbf{Size}} &
 \multicolumn{1}{c}{\textbf{Time}} &
  \multicolumn{1}{c}{\textbf{Time}}

 \\ \hline

SAFE   & 4.4 Kb & 11.86 sec & $<$0.1 msec \\ \hline

GABOR   & 25.5 Kb & 0.53 sec & 0.3 msec  \\
\hline

SIFT   & 2138 Kb & 1.5 sec & 1.1 sec \\
\hline

SIFT 200p & 115.7 Kb & -  &  6.1 msec \\
\hline

SIFT 100p & 58.1 Kb & -  &  2 msec \\
\hline

LBP   & 3.2 Kb & 0.17 sec & $<$0.1 msec \\
\hline

HOG   & 3.6 Kb & 0.13 sec & $<$0.1 msec \\
\hline

\end{tabular}
\end{center}
\caption{Size of the template file and computation times.}
\label{tab:file-size-time}
\end{table}
\normalsize

\begin{figure*}[htb]
\centering
\includegraphics[width=0.92\textwidth]{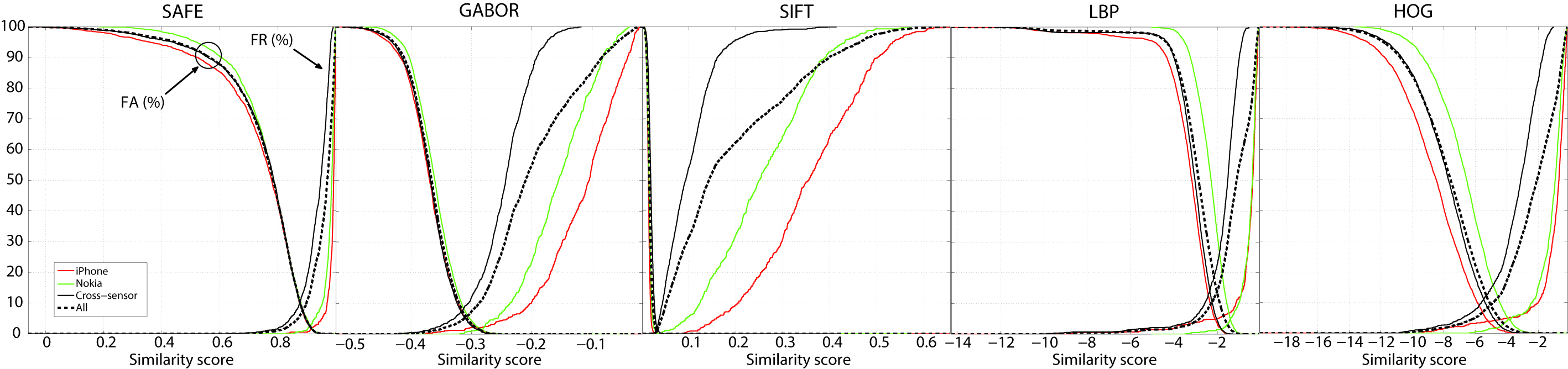}
\caption{Verification results of the individual systems (DET
curves).} \label{fig:DET-indiv}
\end{figure*}

\begin{figure*}[htb]
\centering
\includegraphics[width=0.92\textwidth]{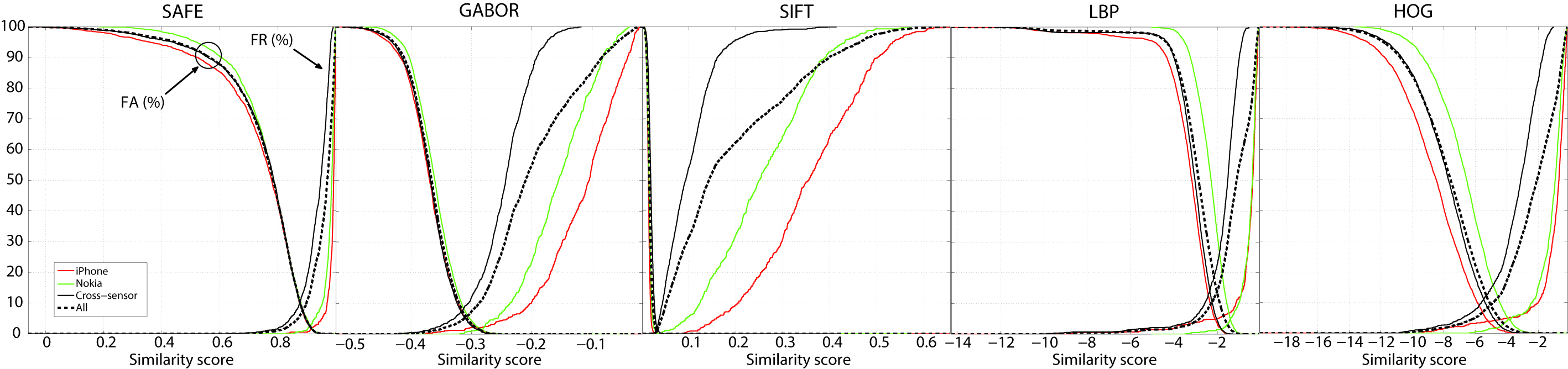}
\caption{Verification results of the individual systems (FA, FR
curves).} \label{fig:FAFR-indiv}
\end{figure*}

\section{Results: Individual Systems}

Performance is reported in Figures~\ref{fig:DET-indiv} and
\ref{fig:FAFR-indiv}. EER values are also given in
Table~\ref{tab:EER-indiv}.
We report: $i$) \emph{same-sensor} comparison; $ii$)
\emph{cross-sensor} comparison; and $iii$) \emph{overall} (pooling
scores of $i$ and $ii$).
We use the SIFT detector as in \cite{[Alonso09]} for iris images,
but here 
it gives $\sim$3000 key-points per image due to a much bigger ROI.
This allows an EER of $\sim$0\% in same-sensor comparisons, but the
template has several MBs and comparison time is $>$1 sec on a laptop
in C++ (Table~\ref{tab:file-size-time}), which may not be feasible
if transferred to devices with limited capabilities.
Comparison time is one of the drawbacks of key-point based systems,
since it is usually needed to compare each key-point of one image
against all key-points of the other.
The other comparators employed have templates of fixed size, thus
comparison is very efficient.
For this reason, we also report results limiting the key-points per
image to 100 and 200 (by changing the threshold to exclude low
contrast points), an approach observed in other studies when image
resolution increases \cite{[Padole12]}.
The SIFT comparator with 100 key-points still has a template and a
comparison time one order of magnitude bigger than some other
systems, but similar performance or even worse.
This indicates that the most $n$ salient key-points of one image
do not necessarily pair fully 
with the most $n$ salient key-points of other image from the same
eye instance, so this limiting approach may not be an efficient
solution either.
%

From Figure~\ref{fig:DET-indiv} and Table~\ref{tab:EER-indiv}, we
observe that even if performance of same-sensor experiments can be
very good, cross-sensor comparison results in a significant
worsening.
There is also correlation between a bigger template
(Table~\ref{tab:file-size-time}) and lower EER
(Table~\ref{tab:EER-indiv}). 
%
It is worth noting too the comparable performance of SAFE w.r.t.
GABOR, with template one fourth in size.
Also, SAFE, LBP and HOG templates are comparable, but performance of
the two latter comparators are worse.
This reflects the discriminative capability of SAFE filters,
although at the expense of a higher extraction time, since
convolution filters are of similar size than the input image
(871$\times$871). However, filter separability could be explored for
faster processing \cite{[Alonso15]}.
%
An interesting observation from Figure~\ref{fig:FAFR-indiv} is that
in the cross-sensor scenario, genuine score distributions (FR curve)
shift significantly towards the impostor distribution (FA), whereas
impostor distributions remain in the same range (at least with SAFE,
GABOR and SIFT). This means that `similarity' between images of the
same instance is reduced when they come from a different
sensor, at least measured by the features employed. 
%
It is also interesting that same-sensor performance is not similar
for each sensor, even if they involve the same eyes, and images have
the same size.
Genuine score distributions are also observed to be in a different
range for each sensor (red and green FR curves of
Figure~\ref{fig:FAFR-indiv}).
We apply local adaptive contrast equalization, but results suggest
however that other device-dependant processing might be of help to
compensate variations in performance \cite{[Santos14]}.


\begin{table}[htb]
\small
\begin{center}
\begin{tabular}{ccccc}


\multicolumn{1}{c}{\textbf{system}} &
 \multicolumn{1}{c}{\textbf{iPhone}} &
 \multicolumn{1}{c}{\textbf{Nokia}} &  \multicolumn{1}{c}{\textbf{cross-sensor}} &
 \multicolumn{1}{c}{\textbf{all}} \\ \hline

SAFE & 1.6\%  & 2.6\% & 10.2\% & 7.5\% \\ \hline

GABOR & 2.1\%  & 1.5\% & 7.3\% & 5.8\% \\
\hline

SIFT & 0\% & 0.1\% & 1.6\%  & 0.8\% \\
\hline

SIFT 200p & 0\% & 0.8\% & 6.6\% & 5\% \\
\hline

SIFT 100p & 0.1\% & 2.1\% & 11.9\% & 8\% \\
\hline

LBP & 4.8\% & 4.9\% & 14.1\% & 13.8\% \\
\hline

HOG & 3.9\% & 4.5\% & 11\% & 9.2\% \\
\hline

\end{tabular}

\end{center}
\caption{Verification results of the individual systems (EER).}
\label{tab:EER-indiv}
\end{table}
\normalsize

\begin{figure}[htb]
\centering
\includegraphics[width=0.48\textwidth]{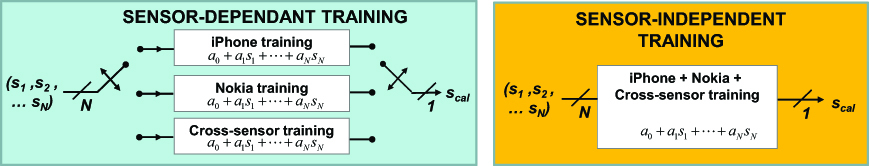}
\caption{Architecture of the two fusion strategies implemented.}
\label{fig:fusion-model}
\end{figure}

\section{Results: Fusion of Periocular Systems}

We carry out fusion experiments using all the available comparators.
Given $N$ comparators which output scores $S$=($s_{1}, s_{2}, ...
s_{N}$) for an input trial, a linear fusion is: $s_{cal} = a_0 + a_1
\cdot s_{1} + ... + a_N \cdot s_{N}$.
Weights $a_{0}, a_{1}, ... a_{N}$ are trained via logistic
regression following a probabilistic Bayesian framework
\cite{[Alonso10]}, in a way that $s_{cal} \simeq \log \left(
{{p\left( {S|\omega _i } \right)}}/{{p\left( {S|\omega _j }
\right)}} \right)$. This is the logarithm of the ratio between the
likelihood that input signals are originated by the same eye
instance (target hypothesis $\omega _i$) or not (non-target
hypothesis $\omega _j$).
%
%
%
An advantage of this approach is that $s_{cal}$ has a probabilistic
value by itself, representing a \emph{degree of support} to any of
the $\omega _i$ and $\omega _j$ hypotheses: if it is higher than 0,
then the support to $\omega _i$ is higher, and vice-versa.
%
%
This trained approach has also shown better performance than simple
fusion rules (like mean or sum) in previous works, and presents
advantages too when signals originate from heterogeneous sources
\cite{[Alonso10]}, as shown next.

\begin{figure}[htb]
\centering
\includegraphics[width=0.4\textwidth]{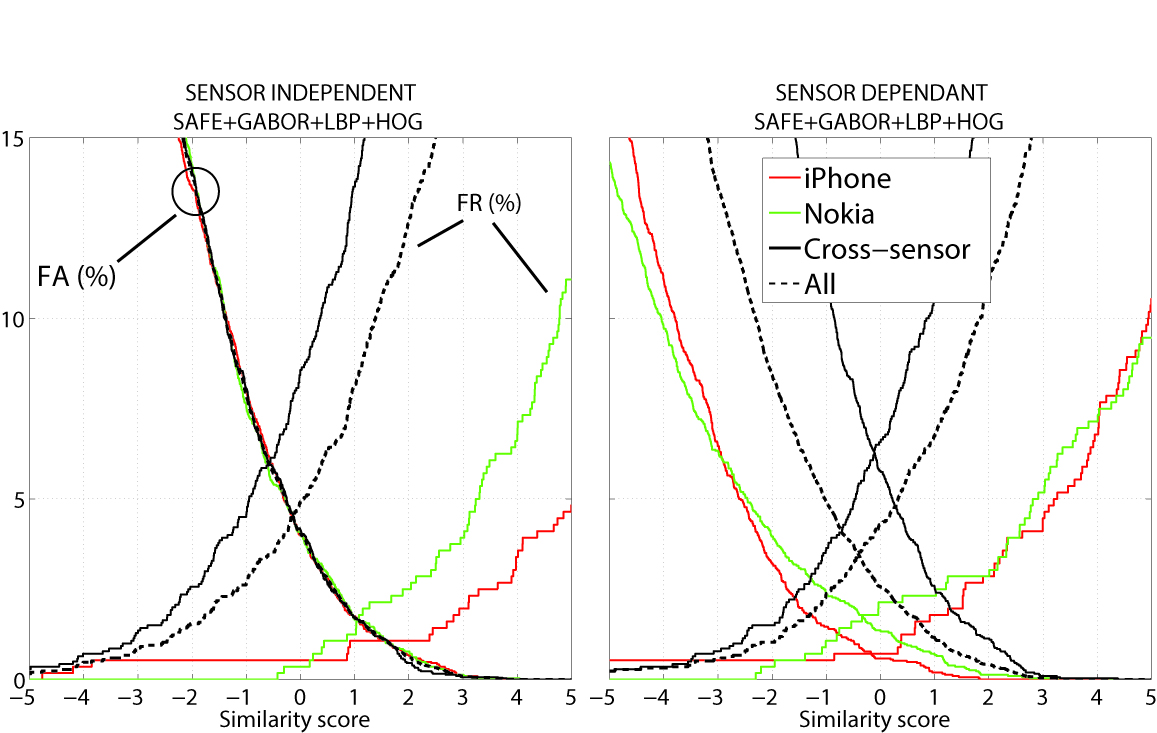}
\caption{Verification results of a fusion example (FA, FR curves).
Left: fusion training is done by pooling same- and cross-sensor
scores; as a result, misalignment between these cases exist. Right:
separate training allows the score distributions to be centered
around a log-likelihood ratio of 0.} \label{fig:fusion-exampleFAFR}
\end{figure}

We evaluate two fusion strategies (Figure~\ref{fig:fusion-model}):
$i$) \emph{sensor-dependant}, with a fusion function trained
separately for \emph{same-sensor} (one per device) and
\emph{cross-sensor} scores; and $ii$) \emph{sensor-independent},
with a unique fusion function trained with same- and cross-sensor
scores together. Case $i$) implies that the device is known, which
is reasonable in operational scenarios, while case $ii$) does not
exploit any knowledge regarding the device used to capture signals.
%
We have tested all possible fusion combinations, with the best
results reported in Table~\ref{tab:EER-results-fusion}. The best
combinations are chosen based on the lowest cross-sensor EER. As it
can be observed, fusion improves cross-sensor performance
significantly, with more than 40\% EER reduction if the local SIFT
comparator is involved; if not (bottom part of
Table~\ref{tab:EER-results-fusion}), cross-sensor performance still
improves 14-18\%.
Regarding the two fusion strategies evaluated, there is no
substantial difference in cross-sensor EER, but performance of
same-sensor tests is equal or better by using sensor-dependant
training. This is because training is done optimally for each
sensor, tailored to differences in the range of similarity scores
observed (Figure~\ref{fig:FAFR-indiv}).
%
%
A further benefit is that same- and cross-sensor score distributions
are aligned after the fusion (Figure~\ref{fig:fusion-exampleFAFR}),
%
providing an elegant and simple solution for handling signals from
different devices, since there is no need of sensor-specific
thresholds.
As a result, global performance as computed by pooling all scores
together (columns `all' in Table~\ref{tab:EER-results-fusion}) is
significantly better as well.

It can also be seen (Table~\ref{tab:EER-results-fusion}) that the
best performance is not necessarily obtained by using all available
systems. Indeed, the highest improvement occurs after the fusion of
two or three systems. Inclusion of more systems produces smaller
improvements (or no improvement at all).
The best performance is given by fusion of only two systems (SAFE,
SIFT), with a cross-sensor EER reduction from 1.6\% to 0.9\% (even
if the cross-sensor performance of SAFE is 10.2\%).
Optimal combinations always involve the SIFT comparator, which also
has the best individual performance (or among the bests when the
number of key-points is limited). The good performance of SIFT is
not jeopardized during the fusion by other comparators with a
performance an order of magnitude worse, but it is complemented to
obtain even better same- and cross-sensor EERs.
This is because in the trained fusion approach employed, the support
of each modality is implicitly weighted by its accuracy. In other
simple fusion methods (such as mean or sum of scores), all
comparator are given the same weight independently of its accuracy.
This is a common problem of these methods, that makes the worst
modalities to yield misleading results more frequently
\cite{[Jain05]}.

A careful look at the best combinations of
Table~\ref{tab:EER-results-fusion} shows that SAFE or GABOR
comparator are always chosen first for the fusion. Together with
SIFT, these are very powerful descriptors that capture different
image features, thus being very complementary too. If we eliminate
SIFT from the equation (bottom of
Table~\ref{tab:EER-results-fusion}), a cross-sensor performance of
$\sim$6\% can be still obtained with the available systems, while
keeping same-sensor performance below 1.5\%.

\begin{table*}[htb]
\tiny
\begin{center}
\begin{tabular}{p{0.07cm}p{0.07cm}p{0.07cm}p{0.07cm}p{0.07cm}p{0.07cm}||c|c|c||c||cp{0.07cm}p{0.07cm}p{0.07cm}p{0.07cm}p{0.07cm}p{0.07cm}||c|c|c||c||}

\multicolumn{21}{c}{\textbf{USING ALL THE SYSTEMS AVAILABLE}} \\

\multicolumn{21}{c}{} \\

\multicolumn{10}{c}{SENSOR-INDEPENDENT TRAINING} & \multicolumn{1}{c}{} & \multicolumn{10}{c}{SENSOR-DEPENDANT TRAINING}  \\

\multicolumn{21}{c}{} \\ \cline{7-10} \cline{18-21}

\multicolumn{1}{p{0.07cm}}{\begin{turn}{90}\textbf{\# systems}
\end{turn}} & \multicolumn{1}{p{0.07cm}}{\begin{turn}{90}\textbf{safe}
\end{turn}} &
\multicolumn{1}{p{0.07cm}}{\begin{turn}{90}\textbf{gabor}
\end{turn}} & \multicolumn{1}{p{0.07cm}}{\begin{turn}{90}\textbf{sift}
\end{turn}} & \multicolumn{1}{p{0.07cm}}{\begin{turn}{90}\textbf{lbp}
\end{turn}} &
\multicolumn{1}{p{0.07cm}||}{\begin{turn}{90}\textbf{hog}
\end{turn}} & iphone & nokia & cross-sensor & all & &
\multicolumn{1}{p{0.07cm}}{\begin{turn}{90}\textbf{\# systems}
\end{turn}} & \multicolumn{1}{p{0.07cm}}{\begin{turn}{90}\textbf{safe}
\end{turn}} &
\multicolumn{1}{p{0.07cm}}{\begin{turn}{90}\textbf{gabor}
\end{turn}} & \multicolumn{1}{p{0.07cm}}{\begin{turn}{90}\textbf{sift}
\end{turn}} & \multicolumn{1}{p{0.07cm}}{\begin{turn}{90}\textbf{lbp}
\end{turn}} &
\multicolumn{1}{p{0.07cm}||}{\begin{turn}{90}\textbf{hog} \end{turn}} & iphone & nokia & cross-sensor & all \\
\cline{1-10} \cline{12-21}

1 & &  & x &  &  & \textbf{0} & 0.1 & 1.6 & 0.8 & & 1 &  &  & x &  &
& \textbf{0} & 0.1  & 1.6  & 0.8
\\ \cline{1-10}
\cline{12-21}

2 & x &  & x &  &  & \textbf{0} (0\%) & \textbf{0} (-100\%) &
\textbf{0.9} (-43.8\%) & \textbf{0.4} (-50\%) & & 2 &  x &  & x &  &
& \textbf{0} (0\%) & \textbf{0} (-100\%) & \textbf{0.9} (-43.8\%) &
\textbf{0.4} (-50\%)
\\
\cline{1-10} \cline{12-21}

3 & x &  & x &  & x & \textbf{0} (0\%) & \textbf{0} (-100\%) &
\textbf{0.9} (-43.8\%) & \textbf{0.4} (-50\%) & & 3 & x &  & x & x &
& \textbf{0} (0\%) & \textbf{0} (-100\%) & \textbf{0.9} (-43.8\%) &
\textbf{0.4} (-50\%)
\\

\cline{1-10} \cline{12-21}

4 & x & x & x &  & x & \textbf{0} (0\%) & \textbf{0} (-100\%) &
\textbf{0.9} (-43.8\%) & \textbf{0.4}
 (-50\%)  & & 4& x & x & x &  & x & \textbf{0} (0\%) & \textbf{0} (-100\%) & \textbf{0.9} (-43.8\%) &
\textbf{0.4} (-50\%) \\
\cline{1-10} \cline{12-21}

5 & x & x & x & x & x & \textbf{0} (0\%) & \textbf{0} (-100\%) &
\textbf{0.9} (-43.8\%) & \textbf{0.4}
 (-50\%)  & & 5 & x & x & x & x & x & \textbf{0} (0\%) & \textbf{0} (-100\%) & \textbf{0.9} (-43.8\%) &
0.8 (+0\%) \\
\cline{1-10} \cline{12-21}

\multicolumn{6}{r}{\textbf{best}} & \multicolumn{1}{c}{\textbf{0}} &
\multicolumn{1}{c}{\textbf{0}} & \multicolumn{1}{c}{\textbf{0.9}} &
\multicolumn{1}{c}{\textbf{0.4}}
   & & \multicolumn{6}{r}{\textbf{best}} &  \multicolumn{1}{c}{\textbf{0}} & \multicolumn{1}{c}{\textbf{0}} & \multicolumn{1}{c}{\textbf{0.9}}  &
\multicolumn{1}{c}{\textbf{0.4}}  \\

\multicolumn{21}{c}{} \\

\multicolumn{21}{c}{} \\

\multicolumn{21}{c}{\textbf{USING LOCAL SIFT COMPARATOR RESTRICTED TO 200 KEY-POINTS PER IMAGE}} \\

\multicolumn{21}{c}{} \\

\multicolumn{10}{c}{SENSOR-INDEPENDENT TRAINING} & \multicolumn{1}{c}{} & \multicolumn{10}{c}{SENSOR-DEPENDANT TRAINING}  \\

\multicolumn{21}{c}{} \\ \cline{7-10} \cline{18-21}

\multicolumn{1}{p{0.07cm}}{\begin{turn}{90}\textbf{\# systems}
\end{turn}} & \multicolumn{1}{p{0.07cm}}{\begin{turn}{90}\textbf{safe} \end{turn}}
& \multicolumn{1}{p{0.07cm}}{\begin{turn}{90}\textbf{gabor}
\end{turn}} & \multicolumn{1}{p{0.07cm}}{\begin{turn}{90}\textbf{sift 200p}
\end{turn}} & \multicolumn{1}{p{0.07cm}}{\begin{turn}{90}\textbf{lbp}
\end{turn}} &
\multicolumn{1}{p{0.07cm}||}{\begin{turn}{90}\textbf{hog}
\end{turn}} & iphone & nokia & cross-sensor & all  & &
\multicolumn{1}{p{0.07cm}}{\begin{turn}{90}\textbf{\# systems}
\end{turn}} & \multicolumn{1}{p{0.07cm}}{\begin{turn}{90}\textbf{safe}
\end{turn}} &
\multicolumn{1}{p{0.07cm}}{\begin{turn}{90}\textbf{gabor}
\end{turn}} & \multicolumn{1}{p{0.07cm}}{\begin{turn}{90}\textbf{sift}
\end{turn}} & \multicolumn{1}{p{0.07cm}}{\begin{turn}{90}\textbf{lbp}
\end{turn}} &
\multicolumn{1}{p{0.07cm}||}{\begin{turn}{90}\textbf{hog} \end{turn}} & iphone & nokia & cross-sensor & all\\
\cline{1-10} \cline{12-21}

1 &  &  & x &  &  & \textbf{0} & 0.8 & 6.6 & 5  & & 1&  &  & x &  &
& \textbf{0}  & 0.8  & 6.6  & 5
\\ \cline{1-10}
\cline{12-21}

2 &  & x & x &  &  & \textbf{0} (0\%) & 0.5 (-37.5\%) & 3.1 (-53\%)
& 2.7 (-46\%)  & & 2 &  & x & x &  &  & \textbf{0} (0\%) & 0.5
(-37.5\%) & 3.1
 (-53\%) & 1.4 (-72\%)
\\ \cline{1-10} \cline{12-21}

3 & x & x & x &  &  & \textbf{0} (0\%) & \textbf{0.4} (-50\%) &
\textbf{2.9} (-56.1\%) & 2.5 (-50\%)  & & 3 &  & x & x &  & x &
\textbf{0} (0\%) & 0.5 (-37.5\%) & \textbf{2.8} (-57.6\%) & 1.4
(-72\%)
 \\

\cline{1-10} \cline{12-21}

4 & x & x & x & x &  & \textbf{0} (0\%) & \textbf{0.4} (-50\%) & 3
(-54.5\%) & 2.3 (-54\%) & & 4& x & x & x &  & x & \textbf{0} (0\%) &
\textbf{0.3} (-62.5\%) & \textbf{2.8} (-57.6\%) & 2.1 (-58\%)

\\ \cline{1-10} \cline{12-21}

5 & x & x & x & x & x & \textbf{0} (0\%) & \textbf{0.4} (-50\%) & 3
(-54.5\%) & \textbf{2.2}
 (-56\%)  & & 5& x & x & x & x & x & \textbf{0} (0\%) & \textbf{0.3} (-62.5\%) &
2.9 (-56.1\%) & \textbf{1.3} (-74\%)
\\
\cline{1-10} \cline{12-21}

\multicolumn{6}{r}{\textbf{best}} & \multicolumn{1}{c}{\textbf{0}} &
\multicolumn{1}{c}{\textbf{0.4}} & \multicolumn{1}{c}{\textbf{2.9}}
& \multicolumn{1}{c}{\textbf{2.2}}
   & & \multicolumn{6}{r}{\textbf{best}} &  \multicolumn{1}{c}{\textbf{0}} & \multicolumn{1}{c}{\textbf{0.3}} & \multicolumn{1}{c}{\textbf{2.8}}  &
\multicolumn{1}{c}{\textbf{1.3}}  \\

\multicolumn{21}{c}{} \\

\multicolumn{21}{c}{} \\

\multicolumn{21}{c}{\textbf{USING LOCAL SIFT COMPARATOR RESTRICTED TO 100 KEY-POINTS PER IMAGE}} \\

\multicolumn{21}{c}{} \\

\multicolumn{10}{c}{SENSOR-INDEPENDENT TRAINING} & \multicolumn{1}{c}{} & \multicolumn{10}{c}{SENSOR-DEPENDANT TRAINING}  \\

\multicolumn{21}{c}{} \\ \cline{7-10} \cline{18-21}

\multicolumn{1}{p{0.07cm}}{\begin{turn}{90}\textbf{\# systems}
\end{turn}} & \multicolumn{1}{p{0.07cm}}{\begin{turn}{90}\textbf{safe}
\end{turn}} &
\multicolumn{1}{p{0.07cm}}{\begin{turn}{90}\textbf{gabor}
\end{turn}} & \multicolumn{1}{p{0.07cm}}{\begin{turn}{90}\textbf{sift 100p}
\end{turn}} & \multicolumn{1}{p{0.07cm}}{\begin{turn}{90}\textbf{lbp}
\end{turn}} &
\multicolumn{1}{p{0.07cm}||}{\begin{turn}{90}\textbf{hog}
\end{turn}} & iphone & nokia & cross-sensor & all  & &
\multicolumn{1}{p{0.07cm}}{\begin{turn}{90}\textbf{\# systems}
\end{turn}} & \multicolumn{1}{p{0.07cm}}{\begin{turn}{90}\textbf{safe}
\end{turn}} &
\multicolumn{1}{p{0.07cm}}{\begin{turn}{90}\textbf{gabor}
\end{turn}} & \multicolumn{1}{p{0.07cm}}{\begin{turn}{90}\textbf{sift}
\end{turn}} & \multicolumn{1}{p{0.07cm}}{\begin{turn}{90}\textbf{lbp}
\end{turn}} &
\multicolumn{1}{p{0.07cm}||}{\begin{turn}{90}\textbf{hog} \end{turn}} & iphone & nokia & cross-sensor & all\\
\cline{1-10} \cline{12-21}

1 &  & x &  &  &  & 0.1 & 1.5 & 7.3 & 5.8  & & 1&  & x &  &  &  &
0.1
  & 1.5  & 7.3  & 5.8
\\ \cline{1-10}
\cline{12-21}

2 &  & x & x &  &  & 0.3 (+200\%) & 1.2 (-20\%) & 4.9 (-32.9\%) &
3.8 (-34.5\%)  & & 2&  & x & x &  &  & \textbf{0} (-100\%) & 1.2
 (-20\%) & 4.9 (-32.9\%) & 2.5
(-56.9\%)
\\ \cline{1-10} \cline{12-21}

3 & x & x & x &  &  & 0.3 (+200\%) & \textbf{0.9} (-40\%) & 4.6
(-37\%) & 3.5
 (-39.7\%)  & & 3& x & x & x &  &  & \textbf{0} (-100\%) & \textbf{0.9} (-40\%)
& 4.7 (-35.6\%) & 2.2 (-62.1\%)
\\

\cline{1-10} \cline{12-21}

4 & x & x & x & x &  & 0.3 (+200\%) & \textbf{0.9} (-40\%) &
\textbf{4.2} (-42.5\%) & \textbf{3.3} (-43.1\%) & & 4& x & x & x & x
&  & \textbf{0} (-100\%) & \textbf{0.9} (-40\%) & \textbf{4.2}
(-42.5\%) & \textbf{2.1} (-63.8\%)

\\ \cline{1-10} \cline{12-21}

5 & x & x & x & x & x & 0.3 (+200\%) & \textbf{0.9} (-40\%) & 4.4
(-39.7\%) & 3.4 (-41.4\%)  & & 5& x & x & x & x & x & \textbf{0}
(-100\%) & \textbf{0.9}
 (-40\%) & 4.3 (-41.1\%) & \textbf{2.1}
(-63.8\%)
\\ \cline{1-10} \cline{12-21}

\multicolumn{6}{r}{\textbf{best}} & \multicolumn{1}{c}{\textbf{0.2}}
& \multicolumn{1}{c}{\textbf{0.9}} &
\multicolumn{1}{c}{\textbf{4.2}} & \multicolumn{1}{c}{\textbf{3.3}}
   & & \multicolumn{6}{r}{\textbf{best}} &  \multicolumn{1}{c}{\textbf{0}} & \multicolumn{1}{c}{\textbf{0.9}} & \multicolumn{1}{c}{\textbf{4.2}}  &
\multicolumn{1}{c}{\textbf{2.1}}  \\

\multicolumn{21}{c}{} \\

\multicolumn{21}{c}{} \\

\multicolumn{21}{c}{\textbf{WITHOUT LOCAL SIFT COMPARATOR}} \\

\multicolumn{21}{c}{} \\

\multicolumn{10}{c}{SENSOR-INDEPENDENT TRAINING} & \multicolumn{1}{c}{} & \multicolumn{10}{c}{SENSOR-DEPENDANT TRAINING}  \\

\multicolumn{21}{c}{} \\ \cline{7-10} \cline{18-21}

\multicolumn{1}{p{0.07cm}}{\begin{turn}{90}\textbf{\# systems}
\end{turn}} & \multicolumn{1}{p{0.07cm}}{\begin{turn}{90}\textbf{safe} \end{turn}}
& \multicolumn{1}{p{0.07cm}}{\begin{turn}{90}\textbf{gabor}
\end{turn}} & \multicolumn{1}{p{0.07cm}}{\begin{turn}{90}\textbf{lbp}
\end{turn}} & \multicolumn{1}{p{0.07cm}}{\begin{turn}{90}\textbf{hog}
\end{turn}} &
\multicolumn{1}{p{0.07cm}||}{\begin{turn}{90}\textbf{} \end{turn}} &
iphone & nokia & cross-sensor & all & &
\multicolumn{1}{p{0.07cm}}{\begin{turn}{90}\textbf{\# systems}
\end{turn}} & \multicolumn{1}{p{0.07cm}}{\begin{turn}{90}\textbf{safe} \end{turn}}
& \multicolumn{1}{p{0.07cm}}{\begin{turn}{90}\textbf{gabor}
\end{turn}} & \multicolumn{1}{p{0.07cm}}{\begin{turn}{90}\textbf{lbp}
\end{turn}} & \multicolumn{1}{p{0.07cm}}{\begin{turn}{90}\textbf{hog}
\end{turn}} &
\multicolumn{1}{p{0.07cm}||}{\begin{turn}{90}\textbf{} \end{turn}} & iphone & nokia & cross-sensor & all\\
\cline{1-10} \cline{12-21}

1 &  & x &  &  &  & 2.1 & \textbf{1.5} & 7.3 & 5.8  & & 1&  & x &  &
&  & 2.1
  & \textbf{1.5}  & 7.3  & 5.8
\\ \cline{1-10}
\cline{12-21}

2 & x & x &  &  &  & 1.6 (0\%) & 1.8 (+20\%) & 6.9 (-5.5\%) & 5.2
 (-10.3\%)  & & 2& x & x &  &  &  & 1.2 (-25\%) & 1.8 (+20\%) &
6.9 (-5.5\%) & 4 (-31\%)
\\
\cline{1-10} \cline{12-21}

3 & x & x & x &  &  & \textbf{1.1} (-31.3\%) & 1.8 (+20\%) & 6.2
(-15.1\%) & 4.8 (-17.2\%) & & 3& x & x & x &  &  & 0.9 (-43.8\%) &
1.8
 (+20\%) & \textbf{6.3} (-13.7\%) & \textbf{3.5}
(-39.7\%)

\\ \cline{1-10} \cline{12-21}

4 & x & x & x & x &  & \textbf{1.1} (-31.3\%) & 1.6 (+6.7\%) &
\textbf{6} (-17.8\%) & \textbf{4.5} (-22.4\%) & & 4& x & x & x & x &
& \textbf{0.7} (-56.3\%) & 1.8 (+20\%) & \textbf{6.3} (-13.7\%) &
\textbf{3.5} (-39.7\%)
\\ \cline{1-10} \cline{12-21}

\multicolumn{6}{r}{\textbf{best}} & \multicolumn{1}{c}{\textbf{1.1}}
& \multicolumn{1}{c}{\textbf{1.5}} & \multicolumn{1}{c}{\textbf{6}}
& \multicolumn{1}{c}{\textbf{4.5}}
   & & \multicolumn{6}{r}{\textbf{best}} &  \multicolumn{1}{c}{\textbf{0.7}} & \multicolumn{1}{c}{\textbf{1.5}} & \multicolumn{1}{c}{\textbf{6.3}}  &
\multicolumn{1}{c}{\textbf{3.5}}  \\

\end{tabular}

\end{center}
\caption{Verification results in terms of EER (in \%) for an
increasing number of fused systems. The best EER achieved for each
case is given, together with the systems involved in the fusion
(best combinations are chosen based on the lowest EER of
cross-sensor experiments). The relative EER variation with respect
to the best individual system is given in brackets. It is also
reported the fusion of 3 systems based on SIFT, LGP and HOG, used as
reference in many periocular studies \cite{[Alonso16]}.}
\label{tab:EER-results-fusion}
\end{table*}
\normalsize

\section{Conclusions}

As biometric technology is increasingly deployed, it will be common
to compare signals from different devices in mismatched conditions.
This issue, known as device interoperability, is known to reduce
performance significantly \cite{[Jain10]}.
We propose the log-likelihood score fusion of several comparators to
improve cross-sensor periocular performance using images from
different smartphones. We evaluate five periocular descriptors of
wide use in the literature.
The database employed has 560 periocular images from two
smartphones.
The fusion scheme is based on linear logistic regression
\cite{[Alonso10]}, in a way that output scores are mapped to
log-likelihood-ratios, thus being in an sensor-independent domain.

Even if the performance when comparing images from the same sensor
can be very good (down to $\sim$0\% with one comparator), an EER
increase of 4 to 10 times is observed when comparing images from
different smartphones.
Score distributions reveal that the `similarity' between images of
the same eyes instance is reduced when they come from a different
sensor, measured by a shift in the genuine scores distribution
towards a range
of smaller similarity values. 
%
An increased intra-class variability is expected in cross-comparison
conditions, due to variability introduced by different imaging
devices \cite{[Alonso10]}.
%
%
For fusion experiments, we consider two strategies
(Figure~\ref{fig:fusion-model}), one that estimates a different
training model for each sensor (sensor-dependent), and another that
trains a single fusion model by pooling both same-sensor and
cross-sensor scores together.
A reduction in cross-sensor performance of more than 40\% can be
achieved with the fusion, with the sensor-dependent strategy
providing additional advantages. For example, since the fusion
function is optimized for each sensor, better performance is
obtained when comparing images from the same device. A further
advantage is that same- and cross-sensor score distributions are
aligned after the fusion, avoiding the use of sensor-specific
decision thresholds and providing significantly better global
performance as well.

Future work includes the use of device-dependant image preprocessing
to compensate variations in image properties \cite{[Santos14]}.
The proposed framework can be applied to comparison of images from
different spectra too \cite{[Jillela14]}.
%
%
In the context of smartphone recognition, where high resolution
images are usual, fusion with the iris modality is another
possibility \cite{[Alonso15a]}. However, it requires segmentation,
which might be an issue if the image quality is not sufficiently
high, which also motivates pursuing the periocular modality, as in
the current study.
We will also validate our methodology using databases not only
limited to two devices, and also including
more extreme variations in camera specifications and imaging conditions.

\small
\section*{Acknowledgment}

This work was done while F. A.-F. was a visiting researcher at the
Norwegian University of Science and Technology in Gj\o{}vik
(Norway), funded by EU COST Action IC1106. F. A.-F. and J. B. also
than the Swedish Research Council, and the CAISR and SIDUS-AIR
programs of the Swedish Knowledge Foundation.



%

\bibliographystyle{IEEEtran}


\end{document}